\documentclass[letterpaper, 10 pt, conference]{ieeeconf}

\IEEEoverridecommandlockouts
\usepackage{cite}
\usepackage{amsmath,amssymb,amsfonts}
\usepackage{algorithm}
\usepackage{algorithmic}
\usepackage{graphicx}
\usepackage{textcomp}
\usepackage{cite}
\usepackage{multirow}
\usepackage{hyperref}
\usepackage[font=small]{caption}
\usepackage{subcaption}
\usepackage{balance}

\usepackage[table,xcdraw]{xcolor}

\makeatletter
    \newcommand{\linebreakand}{%
      \end{@IEEEauthorhalign}
      \hfill\mbox{}\par
      \mbox{}\hfill\begin{@IEEEauthorhalign}
    }
\makeatother

\def\BibTeX{{\rm B\kern-.05em{\sc i\kern-.025em b}\kern-.08em
    T\kern-.1667em\lower.7ex\hbox{E}\kern-.125emX}}
    
\begin{document}

\hbadness=99999 

\title{\LARGE \bf UAV-VLPA*: Vision-Language Mission Planning with Global-Local Optimization for Scalable UAV Route Generation}

\author{ Oleg Sautenkov, Aibek Akhmetkazy$^{*}$, Yasheerah Yaqoot$^{*}$, Muhammad Ahsan Mustafa, \\ Grik Tadevosyan, Artem Lykov, and Dzmitry Tsetserukou
\thanks{The authors are with the Intelligent Space Robotics Laboratory, Center for Digital Engineering, Skolkovo Institute of Science and Technology. 
{\tt \{oleg.sautenkov, aibek.akhmetkazy, yasheerah.yaqoot, ahsan.mustafa, grik.tadevosyan, artem.lykov, d.tsetserukou\}@skoltech.ru}}
\thanks{*These authors contributed equally to this work.}
}

\maketitle


\begin{abstract}
We introduce \textbf{UAV-VLPA*} (Visual-Language-Path-and-Action), a vision-language-based system for autonomous UAV mission planning that generates flight paths directly from natural language instructions and satellite imagery. The system integrates semantic understanding with two key planning components: the Traveling Salesman Problem (TSP) for globally optimized waypoint ordering, and the A* algorithm for obstacle-aware local path refinement.

Tested on the diverse UAV-VLPA-nano-30 benchmark, UAV-VLPA* consistently outperforms human-designed trajectories. Our experiments show that TSP alone reduces the total trajectory length by over \textbf{33.9\%} when compared with the baseline, while the combined TSP + A* setup achieves a \textbf{51.27 km} path, improving on human-generated flight routes by \textbf{18.5\%}. Error analysis across KNN, DTW, and sequential RMSE metrics further confirms the system’s reliability and precision.

Designed for scalable deployment, UAV-VLPA* supports offloading computationally intensive visual-linguistic processing to external servers, enabling lightweight onboard execution. The results demonstrate the system’s ability to autonomously plan efficient and safe UAV missions—without manual intervention—highlighting its potential for real-world applications in complex, large-scale environments.
\end{abstract}
\textbf{\textit{Keywords:}} \textbf{\textit{VLA; VLM; LLM-agents; VLM-agents; UAV; Navigation; Drone; Path Planning.}}

\section{Introduction}
\begin{figure}[t!]
\centering
\includegraphics[width=1\linewidth]{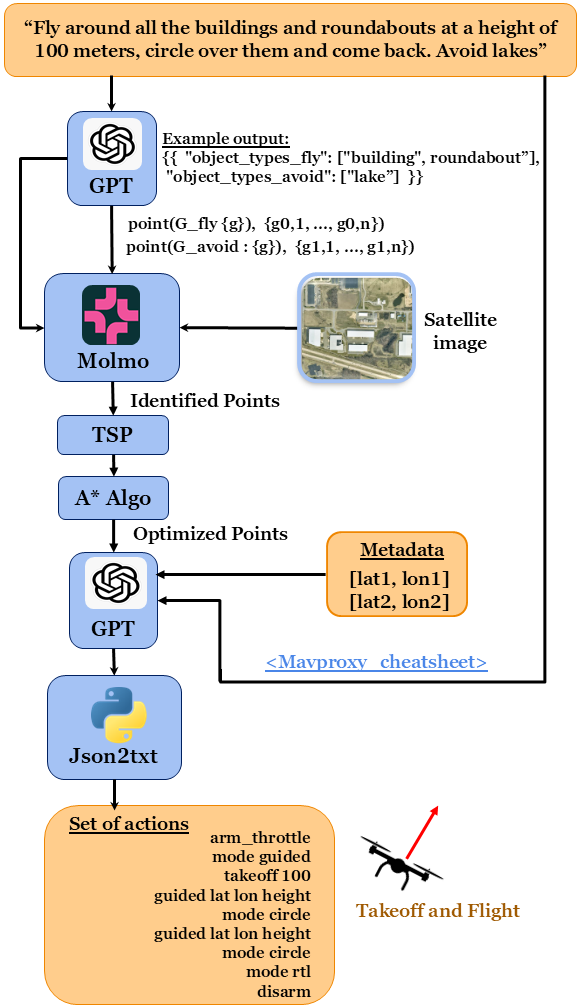} 
\caption{The pipeline of the UAV-VLPA* system.}
\label{software}
\end{figure}

Unmanned aerial vehicles (UAVs) are increasingly used in diverse domains such as surveillance, agriculture, and disaster response~\cite{ai_meets_uav}. As mission complexity grows, the need for intelligent and autonomous path planning becomes critical. Traditional approaches based on manual waypoint setup or pre-programmed routes~\cite{sautenkov2024flightararflightassistance}, \cite{safeswarm} are often inefficient, requiring human expertise and lacking scalability in complex and unknown environments.

Recent work has shown that integrating visual data and natural language commands with navigation modules can enable more autonomous and intuitive UAV control~\cite{lee2024citynavlanguagegoalaerialnavigation}, \cite{ zhong2023safervisionbasedautonomousplanning}, \cite{fan2023aerialvisionanddialognavigation}. Our prior systems, UAV-VLA~\cite{sautenkov2025uavvlavisionlanguageactionlargescale} and UAV-CodeAgents \cite{sautenkov2025uavcodeagentsscalableuavmission}, translates natural language instructions into flight paths using satellite imagery leveraging zero-shot capabilities of powerful models without additional training. Building on this, we present \textbf{UAV-VLPA*}, an improved framework that integrates (1) semantic understanding via vision-language models, (2) global route optimization through the Traveling Salesman Problem (TSP), and (3) local obstacle avoidance using the A* algorithm.

UAV-VLPA* addresses key challenges in UAV navigation by generating efficient, shorter, and safer routes—without requiring manual flight data or expert-tuned instructions. The system autonomously constructs optimal flight plans directly from satellite imagery and user commands, significantly reducing planning time and human effort.

We evaluate our system using the \textbf{UAV-VLPA-nano-30} benchmark~\cite{sautenkov2025uavvlavisionlanguageactionlargescale}, which features diverse real-world environments across the United States, including urban, suburban, rural, and natural scenes. It includes buildings, stadiums, lakes, roads, bridges, and more, captured in daylight during spring and summer. This diversity allows for robust evaluation of route planning capabilities in visually and structurally varied contexts.

\textbf{Our contributions are as follows:}
\begin{itemize}
    \item We propose \textbf{UAV-VLPA*}, combining Vision-Language understanding with TSP and A* for globally optimal and obstacle-aware UAV path planning.
    \item We present a scalable instruction-driven navigation system that eliminates the need for manual flights or environment-specific tuning.
    \item We demonstrate that UAV-VLPA* outperforms human experts and baselines in route efficiency on the UAV-VLPA-nano-30 benchmark.
\end{itemize}

\section{Related Work}
The introduction of Vision Transformers (ViT) \cite{visiontransformerdosovitskiy2021imageworth16x16words},\cite{CLIPradford2021learningtransferablevisualmodels} marked a significant advancement in the development of full-fledged models capable of processing and integrating multiple types of input and output, including text, images, video, and more. Building on this progress, the leading companies introduced models like ChatGPT-4 Omni\cite{openai2024gpt4technicalreport} and DeepSeek-VL \cite{lu2024deepseekvl}, which can reason across audio, vision, and text in real time, enabling seamless multimodal interactions. To address the problem of objects finding in robotics applications, Allen Institute of AI introduced model Molmo, that can point the requested objects on an image\cite{deitke2024molmopixmoopenweights}.

The usage of the transformer-based models allowed the extensive developing of the new methods, benchmarks, and datasets for Vision Language Navigation and Vision Language Action tasks. Firstly, the problem of Aerial Visual Language Navigation was proposed by Liu et al. \cite{liu2023aerialvlnvisionandlanguagenavigationuavs}, where they introduced the Aerial VLN method together with AerialVLN dataset. In \cite{fan2023aerialvisionanddialognavigation} Fan et al. described the simulator and VLDN system, that can support the dialog with an operator during the flight.
Lee et al.\cite{lee2024citynavlanguagegoalaerialnavigation} presented an extended dataset with geographical meta information (streets, squares, boulevards, etc.). The introduction of dataset was paired with the new approach for goal predictor. Zhang et al. 
\cite{EmbodiedCityzhang2024} took a pioneering step by building a universal environment for embodied intelligence in an open city. The agents there can perform both VLA and VLN tasks together online. Gao et al.
\cite{gao2024aerialvisionandlanguagenavigationsemantictopometric} presented a method, where a map was provided as a matrix to the LLM model. In that work was introduced the Semantic Topo Metric Representation (STMR) approach, that allowed to feed the matrix map representation into the Large Language Model. In \cite{wang2024realisticuavvisionlanguagenavigation} Wang et al. presented the benchmark and simulator dubbed OpenUAV platform, which provides realistic environments, flight simulation, and comprehensive algorithmic support. To address the problem of a flight support from the human point of view, Sautenkov et. al \cite{sautenkov2024flightararflightassistance} suggested system to visually enhance the drone control with several video streams and object detection. The system helps the operator to conduct the surveillance tasks, however, doesn't exclude him from the general pipeline. 

Google DeepMind introduced the RT-1 model in their study \cite{brohan2023rt1roboticstransformerrealworld}, wherein the model generates commands for robot operation. The researchers collected an extensive and diverse dataset over several months to train the model. Utilizing this dataset, they developed a transformer-based architecture capable of producing 11-dimensional actions within a discrete action space. Building on the foundation of RT-1, the subsequent RT-2 model \cite{brohan2023rt2visionlanguageactionmodelstransfer} integrates the RT-1 framework with a Visual-Language Model, thereby enabling more advanced multimodal action generation in robotic systems. The work of \cite{kim2024openvlaopensourcevisionlanguageactionmodel},
\cite{open_x_embodiment_rt_x_2023}, and 
\cite{bivla} highlights the potential of transformers and end-to-end neural networks to handle complex Vision-Language-Action (VLA) tasks in real time.

\section{System Overview}
\label{sec:system_overview}
Originally, the vision language model (VLM) provided outputs by identifying point objects on an image in a sequential manner, from left to right and top to bottom, without considering obstacles or their spatial relationships \cite{sautenkov2025uavvlavisionlanguageactionlargescale}. To address these limitations, we propose the following steps.

\subsection{System Workflow}

\begin{itemize}
    \item 
    \textbf{Instruction parsing, Vision-Language Processing, and Waypoint Generation:} We process the instruction to identify and extract target objects for visitation and obstacles for avoidance. 
    \begin{figure}[h]
    \centering
    \includegraphics[width=0.6\linewidth]{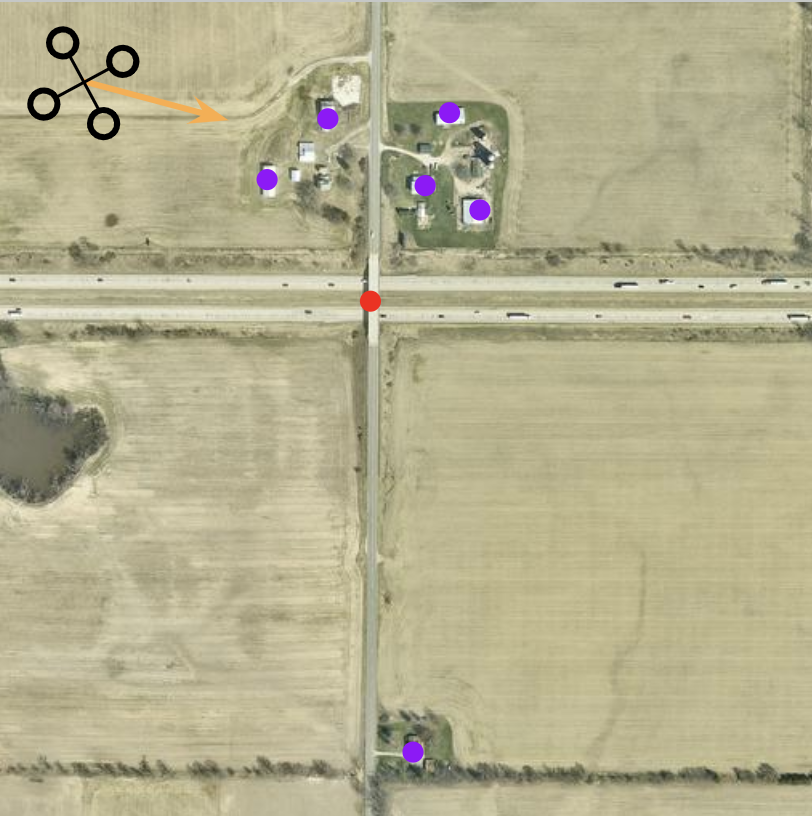}
    \caption{The buildings detected by VLM (purple colored dots)}
    \label{fig:enter-label}
    \end{figure}
    The VLM detects and points obstacles (e.g., buildings) while also interpreting any textual restrictions if they are. As the result, a comprehensive list of waypoints and prohibited regions is produced based on the VLM outputs.

    \item \textbf{TSP Route Construction:} A local search heuristic (2-opt)\cite{dacosta2020learning2optheuristicstraveling} is used to build an initial route that covers all waypoints efficiently.
    \begin{figure}[h]
        \centering
        \includegraphics[width=0.6\linewidth]{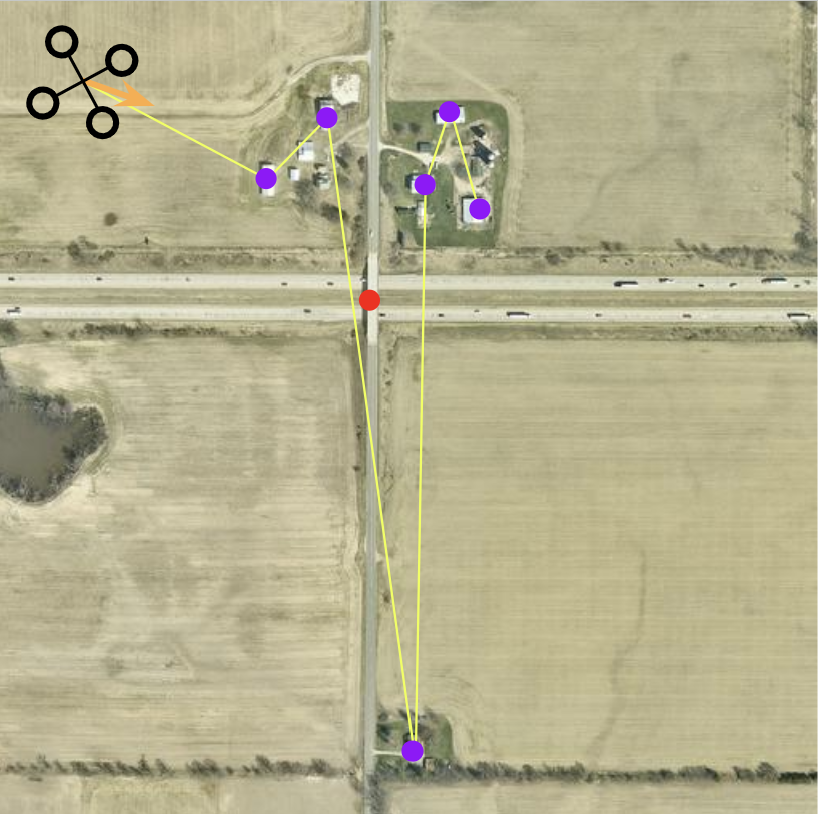}
        \caption{The 2-Opt Local TSP Search path construction through points detected by VLM}
        \label{fig:enter-label}
    \end{figure}

    \item \textbf{A* Path Refinement:} A* then refines each leg of the TSP solution to navigate around obstacles. If necessary, the route is iteratively updated to ensure feasibility.
    \begin{figure}[h]
        \centering
        \includegraphics[width=0.6\linewidth]{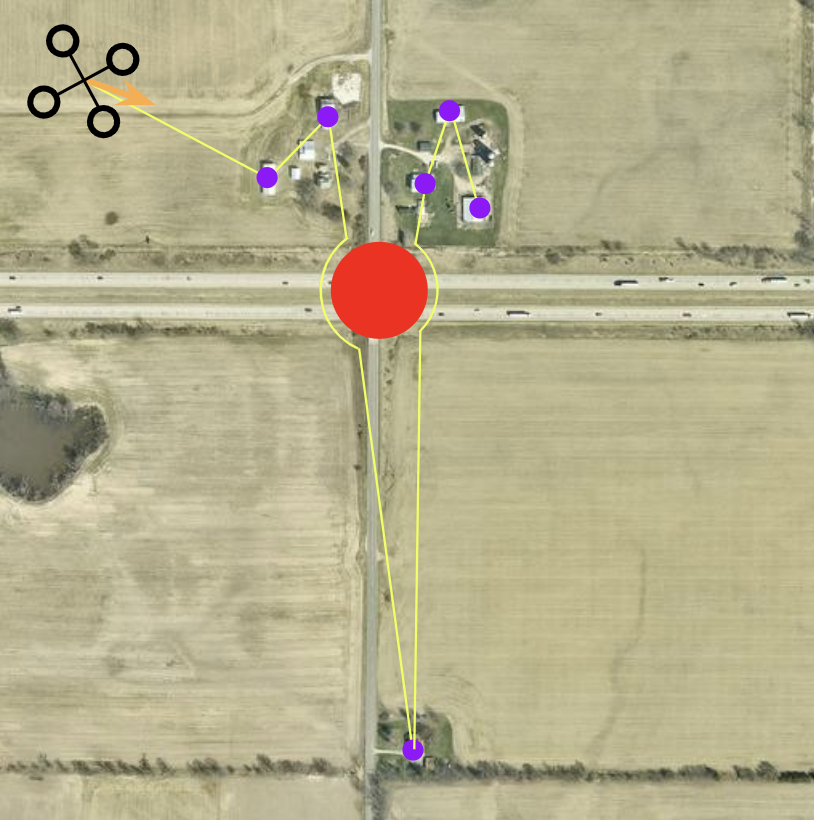}
        \caption{The A* algorithm's route generation considering the obstacle existence through optimal path directions}
        \label{fig:enter-label}
    \end{figure}

    \item \textbf{Final Path Output:} The optimized and obstacle-free route is passed to the post-processing part of the pipeline.
\end{itemize}

\subsection{Traveling Salesman Problem}
The TSP-based component of our system is used to minimize the total travel distance by identifying an optimal or near-optimal tour through a predefined set of waypoints. To achieve this, we implement local search heuristics, specifically the 2-opt scheme \cite{dacosta2020learning2optheuristicstraveling}, which iteratively refines candidate tours to enhance their efficiency. The local search process begins with an initial permutation of waypoints and generates neighboring solutions by systematically swapping pairs of edges. The search continues until no further improvements can be made, at which point the solution converges to a local minimum. It is important to note that this local minimum depends on the specific perturbation scheme employed and may not necessarily coincide with the global minimum. This step ensures that the generated route optimally covers all required waypoints in terms of distance, while detailed obstacle constraints are addressed in subsequent stages. 









\subsection{Path Planning and Obstacle Avoidance}

Since A* algorithm is known as one of the most accurate \cite{FOEAD2021507} and  fastest tools to create a path from one starting point to another goal point, it is chosen as the main algorithm for this work. To efficiently execute the A* algorithm in large-scale environments, the map is discretized into a uniform grid of equally sized cells, where each cell represents a node in the graph. The grid resolution is determined by balancing path fidelity—how accurately the path adheres to real-world terrain and obstacles—and computational efficiency—the number of nodes processed during the search. Considering the balance between smoothness of the trajectory and time-efficiency of the execution, the grid cell size is set to 5 x 5. (discretization step is 5 pixels).


\begin{itemize}
    \item \textbf{Grid Creation:}
    The continuous 2D area is subdivided into $\Delta x \times \Delta y$ cells. For each cell, we check whether it is traversable (i.e., not blocked by an obstacle) based on obstacle data extracted from VLM. 
    
    \item \textbf{Coordinate System Mapping:}
    Each cell is assigned a discrete coordinate pair $(i,j)$ corresponding to its row $(i)$ and column $(j)$ indices.
    
    \item \textbf{Traversability Check:}
    During grid creation, any cell that intersects with a detected obstacle (e.g., a lake) is marked as non-traversable. Such cells are effectively removed from the final graph so that the A* search will not expand them.
\end{itemize}

Another significant reason for selecting this algorithm is its computational efficiency. The algorithm exhibits a time complexity of 
\begin{equation}
\ O(E \log(V))\,
\label{eq:time_complexity}
\end{equation}
 and a space complexity of 
 \begin{equation}
\ O(V)\,
\label{eq:space_complexity}
\end{equation}
where \(E\) represents the number of edges and \(V\) denotes the number of vertices in the constructed graph. These efficiency characteristics make the algorithm highly suitable for large-scale applications, ensuring optimal performance in terms of both computational time and memory usage.

\subsection{Path Planning and Obstacle Avoidance with Traveling Salesman Problem}

To exploit both the global optimization property of TSP and the obstacle-avoidance capability of A*, we introduce a hybrid approach. The TSP solution identifies a minimal traversal order among all waypoints, while A* is integrated on each leg of the TSP tour to avoid obstacles. By combining these methods, we ensure that the route remains both globally efficient and locally safe in the presence of uncertain or densely packed by obstacles environments.

\section{Experimental Setup}

To evaluate the UAV-VLPA* system, we used the natural language instruction: \textit{“Create a flight plan for the quadcopter to fly around each building at a height of 100 m, return to home, and land at the take-off point. Avoid lakes.”} The experiments were conducted on a high-performance workstation equipped with an NVIDIA RTX 4090 GPU (24GB VRAM) and an Intel Core i9-13900K processor. To accommodate hardware limitations, particularly memory constraints, the quantized 4-bit Molmo-7B-O BnB model \cite{molmo_quantized} was utilized.

\section{Experimental Results}

\subsection{Evaluation of TSP Integration on UAV-VLA results}
\label{sec:tsp_only}

To assess the impact of incorporating the Traveling Salesman Problem (TSP) into UAV path optimization, we evaluated the system using the UAV-VLPA-nano-30 benchmark. TSP is a classical combinatorial optimization problem that seeks the shortest possible route visiting a set of locations exactly once and returning to the origin. In the context of UAV mission planning, solving TSP over high-level waypoints (e.g., buildings or mission targets) ensures a globally efficient route, minimizing energy consumption and flight time—critical for resource-constrained UAV operations.

The integration of TSP led to a significant reduction in total flight distance. The generated trajectory covered 51.34 km, compared to 63.89 km in the human-crafted routes reported in the UAV-VLA benchmark \cite{sautenkov2025uavvlavisionlanguageactionlargescale}. Notably, our implementation reduced the original trajectory (77.74 km) by over 33.9\%, confirming the benefits of TSP-based waypoint reordering.

We further evaluated the system using RMSE-based metrics (the same as in the baseline \cite{sautenkov2025uavvlavisionlanguageactionlargescale}: KNN error, DTW error, and Sequential Interpolation error. Results are presented in Table~\ref{table:error_metrics_tsp}:

\begin{table}[h!]
\caption{\textsc{RMSE Metrics for UAV-VLA System with TSP Integration}}
\begin{center}
\begin{tabular}{|c|c|c|c|}
\hline
\rowcolor[HTML]{FFE972} 
\textbf{Metric (RMSE)} & \textbf{KNN (m)} & \textbf{DTW (m)} & \textbf{Sequential (m)} \\
\hline
Mean   & 45.16   & 259.07  & 354.46   \\
\hline
Median & 26.89   & 235.48  & 308.91   \\
\hline
Max    & 336.22  & 719.48  & 734.32   \\
\hline
\end{tabular}
\label{table:error_metrics_tsp}
\end{center}
\end{table}

\begin{figure}[h!]
    \centering
    \includegraphics[width=1.0\linewidth]{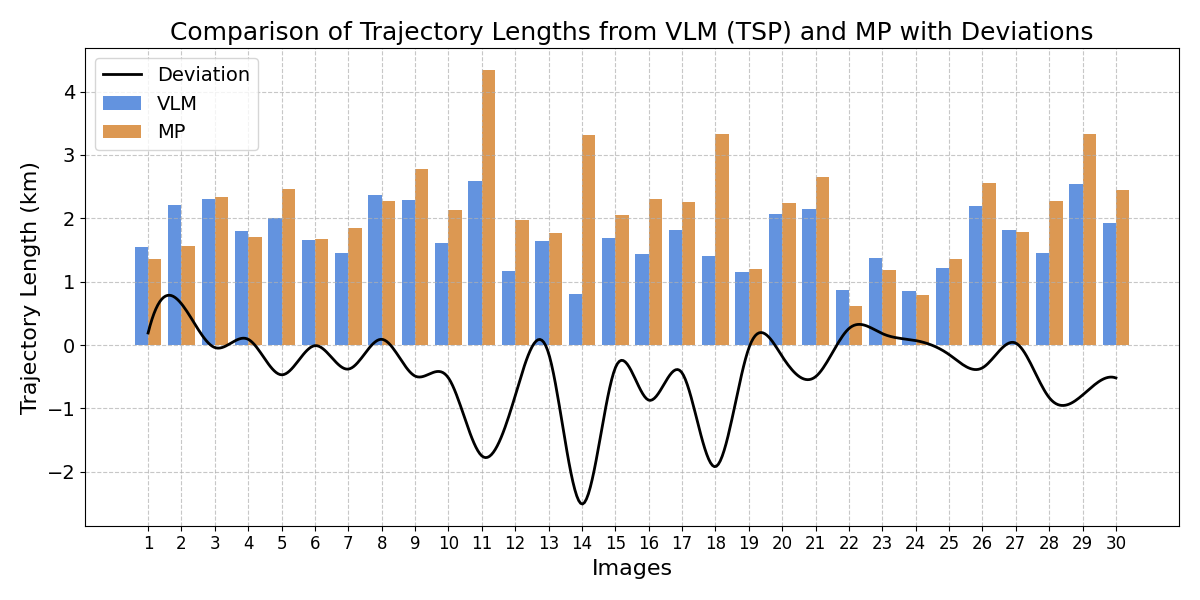}
    \caption{Trajectory length comparison: UAV-VLA with TSP vs. human expert.}
    \label{fig:traj_length_tsp}
\end{figure}

\begin{figure}[h!]
    \centering
    \includegraphics[width=0.7\linewidth]{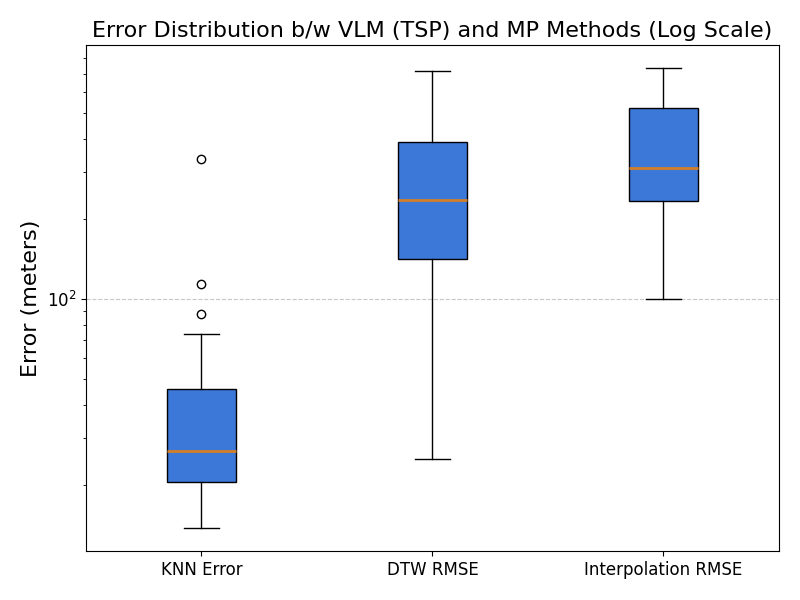}
    \caption{Error distribution of UAV-VLA with TSP against ground truth.}
    \label{fig:errors_tsp}
\end{figure}

Figures~\ref{fig:traj_length_tsp} and \ref{fig:errors_tsp} visualize the comparison in trajectory length and error metrics, respectively. TSP-enhanced flight plans consistently outperformed human-designed ones in terms of efficiency while maintaining competitive accuracy across trajectory alignment metrics.

\subsection{Evaluation of A* Path Planning}

In this and subsequent experiments, the human-generated trajectory was updated to reflect the new constraint of avoiding lakes, which slightly altered both the total trajectory length and mission construction time.

Using only the A* algorithm for low-level path generation—without TSP for waypoint reordering—the system produced a trajectory of 76.84 km. This is notably longer than the updated human-planned route of 62.95 km. The increase suggests that while A* is effective at generating obstacle-avoiding paths, its sequential application without global optimization (e.g., TSP) leads to inefficiencies in overall mission planning.

\begin{figure}[h!]
    \centering
    \includegraphics[width=1.0\linewidth]{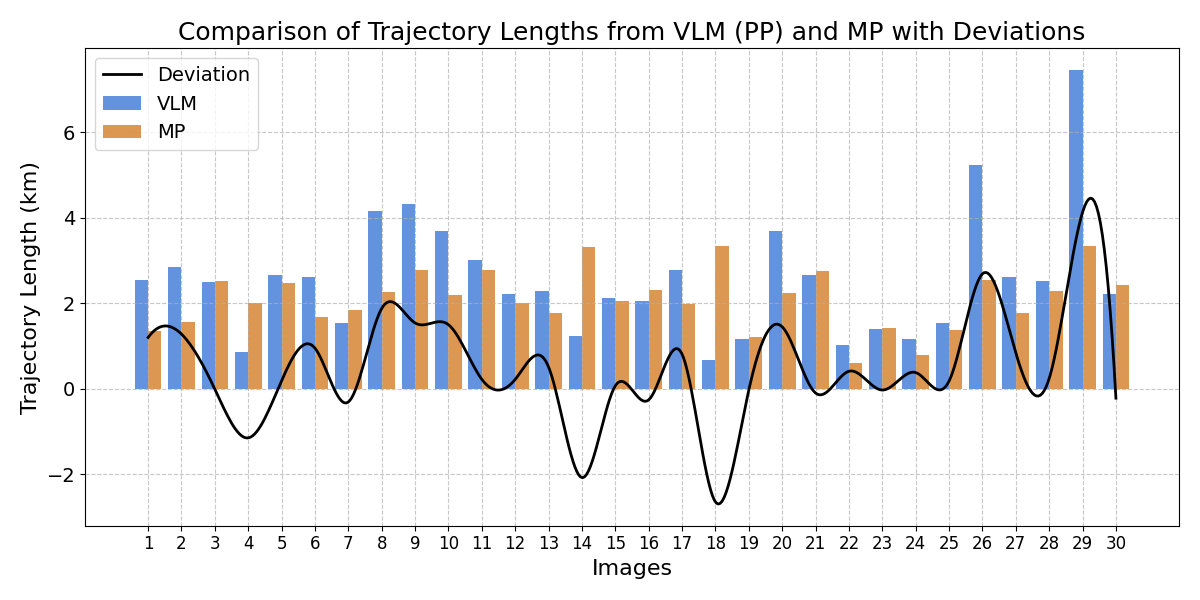}
    \caption{Trajectory length comparison: UAV-VLA with A* vs. human expert.}
    \label{fig:traj_length_pp}
\end{figure}

Figure~\ref{fig:traj_length_pp} shows the comparative trajectory lengths, and Table~\ref{table:error_metrics_pp} presents the corresponding RMSE error metrics.

\begin{table}[htbp]
\caption{\textsc{RMSE Metrics for UAV-VLA System with A* Path Planning}}
\begin{center}
\begin{tabular}{|c|c|c|c|}
\hline
\rowcolor[HTML]{FFE972} 
\textbf{Metric (RMSE)} & \textbf{KNN (m)} & \textbf{DTW (m)} & \textbf{Sequential (m)} \\
\hline
Mean   & 70.15   & 262.68  & 403.31   \\
\hline
Median & 53.65   & 267.01  & 366.28   \\
\hline
Max    & 337.36  & 534.32  & 730.03   \\
\hline
\end{tabular}
\label{table:error_metrics_pp}
\end{center}
\end{table}

\begin{figure}[h!]
    \centering
    \includegraphics[width=0.7\linewidth]{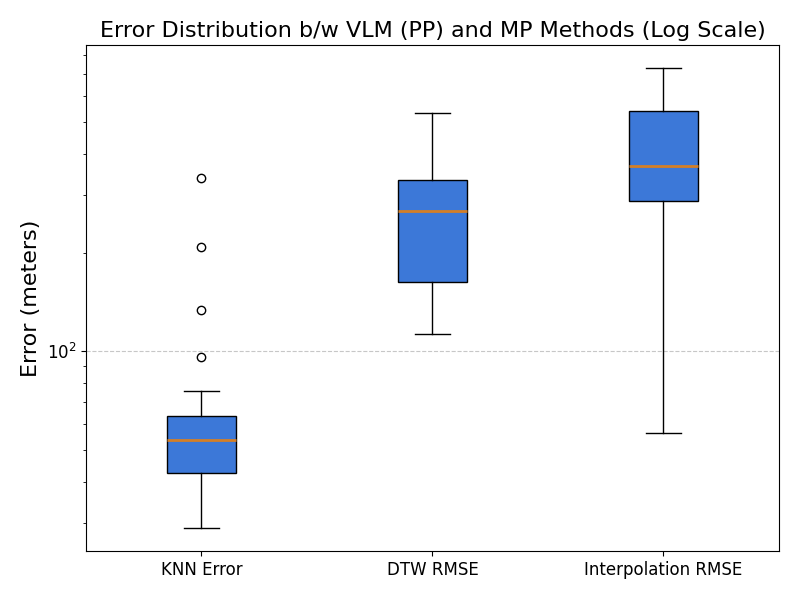}
    \caption{Error distribution of UAV-VLA with A* path planning.}
    \label{fig:errors_pp}
\end{figure}

\begin{figure}[h!]
\centering
\begin{subfigure}{.24\textwidth}
  \centering
  \includegraphics[width=.9\linewidth]{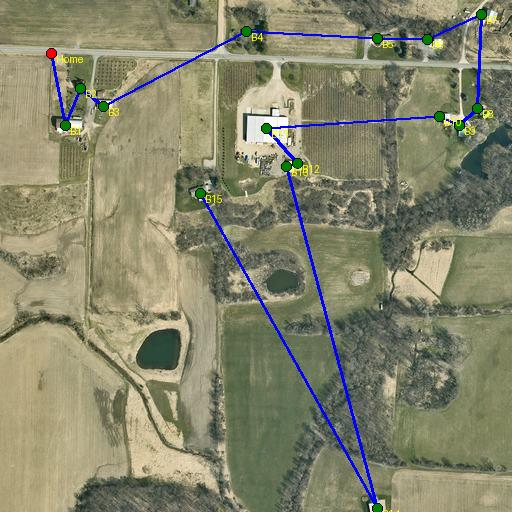}
  \caption{Human-made}
  \label{fig:sub1}
\end{subfigure}%
\begin{subfigure}{.24\textwidth}
  \centering
  \includegraphics[width=.9\linewidth]{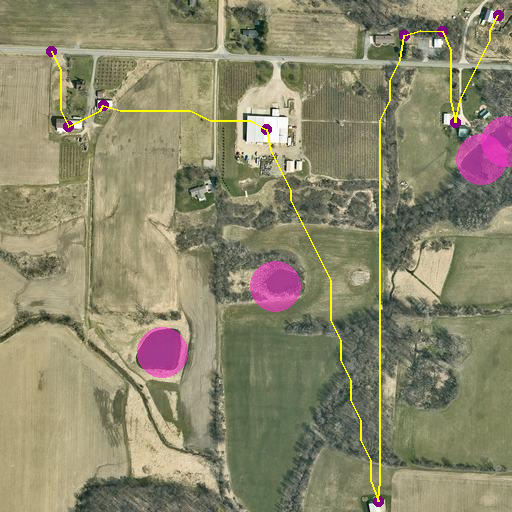}
  \caption{UAV-VLA with A*}
  \label{fig:sub2}
\end{subfigure}
\caption{Flight plans comparison: (a) human expert vs. (b) UAV-VLA with A* path planning.}
\label{fig:example_pp}
\end{figure}

\subsection{Combined TSP and A* Path Planning: UAV-VLPA* System}

When combining both TSP (for waypoint reordering) and A* (for fine-grained obstacle avoidance), the UAV-VLPA* system achieved the best performance. The resulting trajectory was 51.27 km, outperforming the updated human reference trajectory of 62.95 km by 18.5\%.

This result highlights the complementary strengths of the two algorithms: TSP provides a globally optimal waypoint visitation order, while A* ensures feasible and safe traversal between those waypoints.

\begin{figure}[h!]
    \centering
    \includegraphics[width=1.0\linewidth]{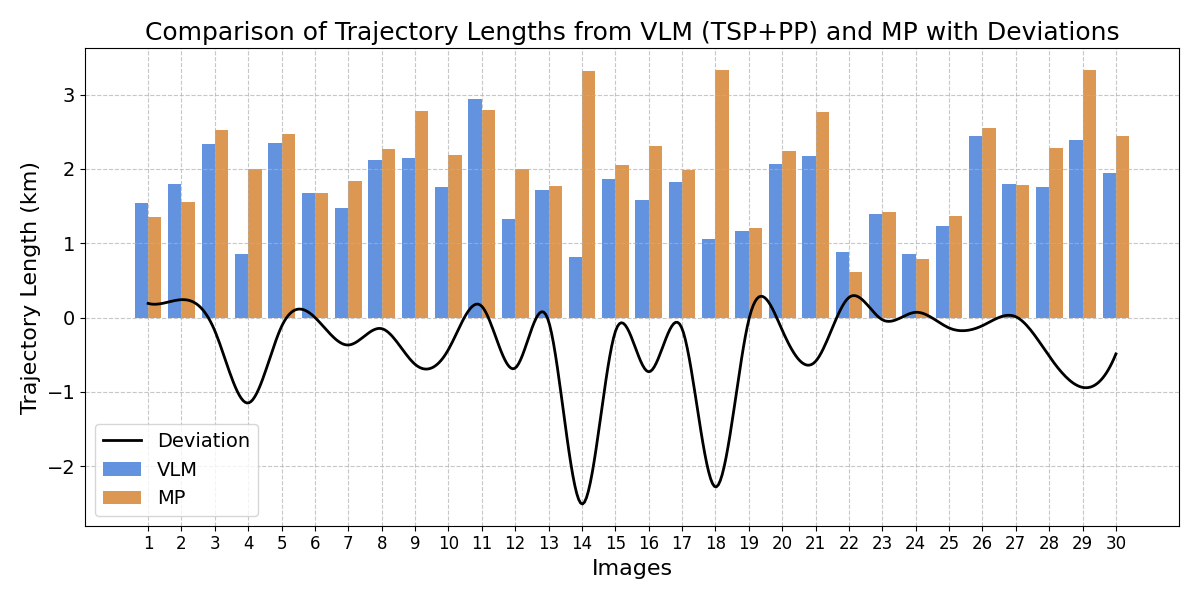}
    \caption{Trajectory length comparison: UAV-VLPA* system vs. human expert.}
    \label{fig:traj_length_tsp_pp}
\end{figure}

\begin{table}[h!]
\caption{\textsc{RMSE Metrics for UAV-VLPA* System (TSP + A*)}}
\begin{center}
\begin{tabular}{|c|c|c|c|}
\hline
\rowcolor[HTML]{FFE972} 
\textbf{Metric (RMSE)} & \textbf{KNN (m)} & \textbf{DTW (m)} & \textbf{Sequential (m)} \\
\hline
Mean   & 69.19   & 223.94  & 348.24   \\
\hline
Median & 54.44   & 170.95  & 291.86   \\
\hline
Max    & 337.89  & 476.07  & 723.49   \\
\hline
\end{tabular}
\label{table:error_metrics_tsp_pp}
\end{center}
\end{table}

\begin{figure}[h!]
    \centering
    \includegraphics[width=0.7\linewidth]{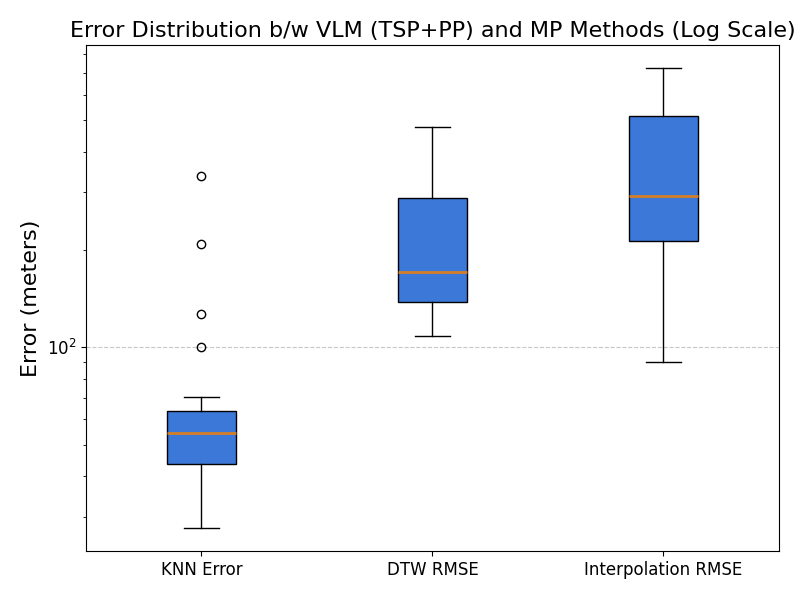}
    \caption{Error distribution of UAV-VLPA* system with TSP and A* integration.}
    \label{fig:errors_tsp_pp}
\end{figure}

\begin{figure}[h!]
\centering
\begin{subfigure}{.24\textwidth}
  \centering
  \includegraphics[width=.9\linewidth]{figures/hm.png}
  \caption{Human-made}
  \label{fig:sub1}
\end{subfigure}%
\begin{subfigure}{.24\textwidth}
  \centering
  \includegraphics[width=.9\linewidth]{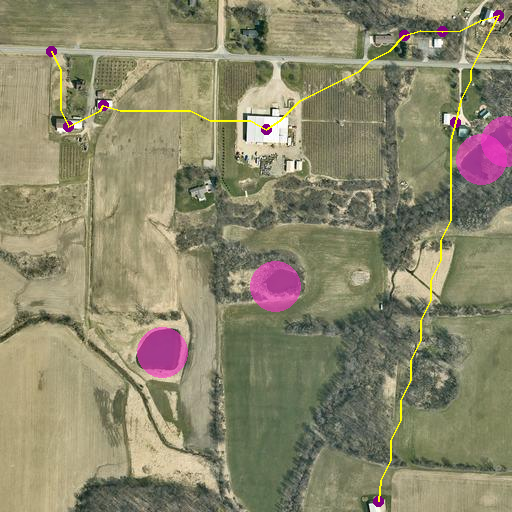}
  \caption{UAV-VLPA*}
  \label{fig:sub2}
\end{subfigure}
\caption{Flight plans comparison: (a) human expert vs. (b) the UAV-VLPA* system (TSP + A*).}
\label{fig:example_tsp_pp}
\end{figure}

In all of the given approaches, the total execution time of the whole pipeline took on HPC no more than 3 minutes, whereas human-made trajectory constructions took in average 35 minutes.




\section{Conclusion}
In this paper, we introduced \textbf{UAV-VLPA*}, a vision-language-based UAV mission planning framework that integrates natural language understanding, satellite imagery, TSP-based global optimization, and A* obstacle-aware path planning. Our goal was to generate efficient, safe, and scalable UAV flight plans directly from user instructions—without requiring manual flight demonstrations or labor-intensive path design.

We evaluated our system using the \textbf{UAV-VLPA-nano-30} benchmark, which features a diverse set of environments across the United States. Through extensive experiments, we demonstrated the effectiveness of each component (TSP, A*, and their combination) in improving route efficiency and mission execution quality.

\textbf{Key findings of this research are:}
\begin{itemize}
    \item \textbf{TSP Integration on the results of the baseline UAV-VLA \cite{sautenkov2025uavvlavisionlanguageactionlargescale}:} Reduced trajectory length by 33.9\% compared to the original baseline, outperforming both human-made and previous system-generated routes.
    \item \textbf{A* Path Planning:} Enabled obstacle avoidance in local path segments but resulted in longer global paths without TSP, showing the importance of combining global and local planning.
    \item \textbf{UAV-VLPA* (TSP + A*):} Achieved the best performance with a trajectory length of 51.27 km, improving upon the human reference path by 18.5\%, while maintaining low error across multiple trajectory alignment metrics.
    \item \textbf{Scalability and Generalization:} The system successfully handled varied geographic regions and structural layouts using open satellite imagery and flexible language input, with no environment-specific tuning.
\end{itemize}

Overall, UAV-VLPA* represents a step forward in intelligent UAV mission generation, offering a practical and scalable solution for natural language-driven flight planning in complex real-world scenarios. Future work will explore integrating 3D terrain data, dynamic obstacle handling, and real-time adaptation to further enhance mission robustness.


\section{Future Work}
The integration of natural language-driven global path planning marks a promising direction for advancing autonomous aerial systems. By combining open-access satellite imagery with classical optimization methods like the Traveling Salesman Problem (TSP) and A* algorithm, our current system lays the groundwork for efficient mission generation in complex and large-scale environments.

Future improvements will focus on addressing real-time operational challenges. Specifically, we aim to:
\begin{itemize}
    \item Enhance computational efficiency for real-time deployment by optimizing the system’s inference pipeline and reducing latency in path generation.
    \item Support dynamic environmental adaptation by integrating up-to-date satellite or UAV-captured visual data, enabling reactive re-planning in response to evolving conditions such as temporary obstacles or weather changes.
    \item Deploy real-world flight experiments where onboard UAV systems will communicate with a remote server. In this setup, high-compute modules (Vision-Language Model and Large Language Model) will be offloaded to the server, while onboard components will handle navigation and control, ensuring lightweight execution and real-time responsiveness.
    \item Expand to 3D environments, incorporating terrain elevation data and urban structure heights to enable more realistic and safe UAV navigation in dense or uneven terrains.
\end{itemize}
By addressing these aspects, future iterations of UAV-VLPA* will be better positioned for real-world deployment, with improved robustness, scalability, and mission execution in dynamic, large-scale aerial scenarios.

\balance

\bibliographystyle{IEEEtran}
\bibliography{ref}

\end{document}